# Constructing Hierarchical Image-tags Bimodal Representations for Word Tags Alternative Choice

**Fangxiang Feng**     f.fangxiang@gmail.com
**Ruifan Li**     rfli@bupt.edu.cn
**Xiaojie Wang**     xjwang@bupt.edu.cn

Engineering Research Center of Information Networks, Ministry of Education,
School of Computer Science, Beijing University of Posts and Telecommunications, Beijing, 100876 China

## Abstract

This paper describes our solution to the multi-modal learning challenge of ICML. This solution comprises constructing three-level representations in three consecutive stages and choosing correct tag words with a data-specific strategy. Firstly, we use typical methods to obtain level-1 representations. Each image is represented using MPEG-7 and gist descriptors with additional features released by the contest organizers. And the corresponding word tags are represented by bag-of-words model with a dictionary of 4000 words. Secondly, we learn the level-2 representations using two stacked RBMs for each modality. Thirdly, we propose a bimodal auto-encoder to learn the similarities/dissimilarities between the pairwise image-tags as level-3 representations. Finally, during the test phase, based on one observation of the dataset, we come up with a data-specific strategy to choose the correct tag words leading to a leap of an improved overall performance. Our final average accuracy on the private test set is 100%, which ranks the first place in this challenge.

## 1. Introduction

The multi-modal learning challenge of ICML 2013 aims at developing a predictive system for word tags using bimodal data: images and texts. Specifically, the data used in this contest contains two groups: the Small ESP Game Dataset (von Ahn & Dabbish, 2004) for training created by Luis von Ahn and the manually labeled dataset for test by Ian Goodfellow. In the rest of this paper, we refer to these two datasets as ESP and GF, respectively. The ESP consists of 100,000 labeled images with tags. The GF consists of 1000 test examples come in triples: an image, and two annotations, i.e. a correct description and an incorrect one. The GF is further evenly divided into public test set and private test set. The performance of the predictions is evaluated based on the accuracy at predicting which of the two descriptions fits the image better. Below we describe some important properties of the these two datasets:

- Some statistical differences exist between these two datasets. The images in ESP have a variety of sizes, while the test images are 300-pixel long on the larger dimension.

- For each image in GF, the incorrect description is always the correct description of one other test image.

This paper describes our solution to handle the above challenges. Our approach treats an image and its tag words as a pair of data for the same hidden object and endeavors to model the similar representations between these two types of descriptions.

The following sections describe our solution in detail. The architecture of our approach is outlined in section 2. Then the three consecutive stages for constructing representations are described in sections 3, 4, and 5, successively. Section 6 introduces our strategies for choosing word tags. Section 7 shows our experimental results.

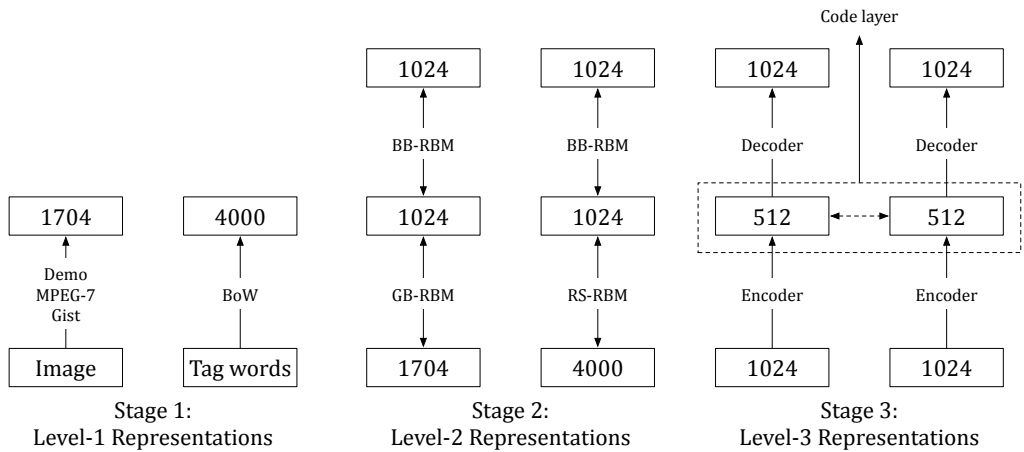

Figure 1. The system architecture of our solution.

## 2. System Architecture

The main idea of our solution is that we endeavor to construct hierarchical representations of bimodal data for choosing the word tags. In the training phase, we represent the data using three consecutive stages. In the first stage, the low-level representations for these two types of data are obtained respectively. For images, the features released by the contest organizer, extracted by four descriptors in MPEG-7, and images gist features are combined to form the level-1 representations. For tag words, the typical bag-of-words model is used for level-1 representations. In the second stage, the level-1 representations for image and tag words are distilled to form the level-2 representation using two stacked Restricted Boltzmann Machines (RBMs), respectively. In the third stage, we propose a quasi-Siamese auto-encoder for learning the level-3 similar representations of these bimodal data. The details of this network are given in section 5.

The architecture of our solution is shown in Figure 1. In this figure, three sub-figures show the three stages for representation constructions. The digits in the boxes are the numbers of neurons used for each layer except the two boxes for images and tag words. The detailed description of each stage is presented in the following sections.

In the test phase, a new pair of data (an image and one of its tags) is given to the three-stage modules, which can obtain the similarity/dissimilarity between the pair of data. By comparing the dissimilarity of two tags with the image, the tag with smaller dissimilarity is chosen as the alternative. For this task, a particular strategy is utilized to improve the accuracy.

## 3. Obtaining Level 1 Representations

Because of the bimodal nature of this competition, we represent our input data from two perspectives: image and text. For image representation we adopt three types of features: the features from contest organizer, the MPEG-7, and gist descriptors. The contest organizer released some extracted image features with 816 dimensions. We remove the invalid 408 all-zero dimensions, reducing the size of features from 816 to 408.

Besides that, we use MPEG-7 and gist descriptors. One part of MPEG-7 is a standard for visual descriptors. We use four different visual descriptors defined in MPEG-7 for image representations: Color Layout (CL), Color Structure (CS), Edge Histogram (EH), and Scalable Color (SC). CL is based on spatial distribution of colors. It is obtained applying the DCT transformation. We used 192 coefficients. CS is based on color distribution and local spatial structure of the color. We used the 256 coefficients form. EH is based on spatial distribution of edges (fixed 80 coefficients). SC is based on the color histogram in HSV color space encoded by a Haar transform. We used the form of 256 coefficients. The software module based on the MPEG-7 Reference Software, available at http://www.cs.bilkent.edu.tr/~bilmdg/bilvideo-7/Software.html, permits obtaining all four different descriptors. Thus, we extract the features of MPEG-7 with the size of 784.

Gist represents the dominant spatial structure of a scene by a set of perceptual dimensions, including naturalness, openness, roughness, expansion, and ruggedness. These perceptual dimensions can be estimated using spectral and coarsely localized information. In our experiments, we use the package from http://people.csail.mit.edu/torralba/code/spatialenvelope/ for image gist descriptor. From all the three groups of features, each image can be represented as a vector of 1704 dimensions. Among them, the first 408 dimensions stand for features provided by the organizer, the middle 784 dimensions for MPEG-7, and the last 512 dimensions for gist features.

For tags representation we use bag-of-words model. A dictionary of 4000 high-frequency words is built from all the tag words of ESP. Then, each word in one image tag can be represented as a multinomial variable. Conveniently, the 1-of-4000 coding scheme is adopted. Thus, each tag can be represented as a vector with 4000 1/0 elements, in which each element stands for whether the tag word is in the dictionary or not. For tag words of an image in the dictionary, they are encoded as 1s, and vice versa.

## 4. Learning Level 2 Representations

In the second stage, we use RBMs to construct the level-2 representations. Those level-1 representations obtained in the first stage for images and tag words have different properties. That is, the level-1 representations of images have real values and those of tag words have multiple 1/0 values. We model these two types of data by different variants of RBMs: Gaussian-Bernoulli RBM and Replicated softmax, respectively. Below, we describe some key points of those learning machines.

### 4.1. Restricted Boltzmann Machines

RBM (Smolensky, 1986) is an undirected graphical model with stochastic binary units in visible layer and hidden layer but without connections between units within these two layers. Given that there are $n$ visible units $\mathbf{v}$ and $m$ hidden units $\mathbf{h}$, and each unit is distributed by Bernoulli distribution with logistic activation function $\sigma(x) = 1/(1 + \exp(-x))$, we then define a joint probabilistic distribution of visible units $\mathbf{v}$ and hidden units $\mathbf{h}$

$$p(\mathbf{v}, \mathbf{h}) = \frac{1}{Z} \exp\left(-E(\mathbf{v}, \mathbf{h})\right) \quad (1)$$

in which, $Z$ is the normalization constant and $E(\mathbf{v}, \mathbf{h})$ is the energy function defined by the configurations of all the units as

$$E(\mathbf{v}, \mathbf{h}) = -\sum_{i=1}^{n}\sum_{j=1}^{m} w_{ij} v_i h_j - \sum_{i=1}^{n} b_i v_i - \sum_{j=1}^{m} c_j h_j \quad (2)$$

By maximizing the log-likelihood of input data, we can learn the parameters of an RBM. This can be achieved by gradient decent. And the weights updates using

$$\Delta w_{ij} = \epsilon \cdot \frac{\partial \log p(v)}{\partial w_{ij}} = \epsilon \cdot \left(\langle v_i h_j \rangle_{data} - \langle v_i h_j \rangle_{model}\right) \quad (3)$$

in which, $\epsilon$ is the learning rate, and $\langle \cdot \rangle$ is the operator of expectation with the corresponding distribution denoted by the subscript. The activation of visible units and hidden units can be infered by the following two equations

$$p(h_j = 1|\mathbf{v}) = \sigma\left(c_j + \sum_{i=1}^{n} w_{ij} v_i\right) \quad (4)$$

$$p(v_i = 1|\mathbf{h}) = \sigma\left(b_i + \sum_{j=1}^{n} w_{ij} h_j\right) \quad (5)$$

in which, $\sigma(\cdot)$ is the logistic activation function.

### 4.2. Modeling Real-valued Data

We model the real-valued data using Gaussian RBM, which is an extension of the binary RBM replacing the Bernoulli distribution with Gaussian distribution for the visible data (Welling et al., 2004). The energy function of different configurations of visible units and hidden ones are

$$E(\mathbf{v}, \mathbf{h}) = -\sum_{i=1}^{n}\sum_{j=1}^{m} w_{ij} \frac{v_i}{\sigma_i} h_j + \sum_{i=1}^{n} \frac{(v_i - b_i)^2}{2\sigma^2} - \sum_{j=1}^{m} c_j h_j \quad (6)$$

The gradient of the log-likelihood function is:

$$\frac{\partial \log p(\mathbf{v})}{\partial w_{ij}} = \left\langle \frac{v_i}{\sigma_i} h_j \right\rangle_{data} - \left\langle \frac{v_i}{\sigma_i} h_j \right\rangle_{model} \quad (7)$$

Usually, we set the variances $\sigma^2 = 1$ for all visible units.

### 4.3. Modeling Count Data

For the count data, we use Replicated Softmax Model (Salakhutdinov & Hinton, 2009) for modeling this sparse vectors. The energy function of the sate configurations is defined as follows

$$E(\mathbf{v}, \mathbf{h}) = -\sum_{i=1}^{n}\sum_{j=1}^{m} w_{ij} v_i h_j - \sum_{i=1}^{n} b_i v_i - M \sum_{j=1}^{m} c_j h_j \quad (8)$$

where $M$ is the total number of words in a document. Note that this replicated softmax model can be interpreted as an RBM model that uses a single visible multinomial unit with support $1, \ldots, K$ which is sampled $M$ times. That is, for each document we create

Constructing Hierarchical Image-tags Bimodal Representations for Word Tags Alternative Choice

a separate RBM with as many softmax units as there are words in the document.

We can efficiently learn all the model by using the Contrastive Divergence approximation (CD) (Hinton, 2002).

In our solution, for each modality we stack two RBMs to learn the level-2 representations. These two-layer stacked RBMs can be trained by greedy layer-wise method (Hinton et al., 2006; Bengio et al., 2007).

## 5. Learning Level 3 Representations

In the third stage, we propose a quasi-Siamese auto-encoder for bimodal representations. The Siamese architecture of neural networks is originally proposed for signature verification (Bromley et al., 1993). The network takes a pair of signature patterns either from the same person or not as inputs. The loss function is simultaneously optimized by minimizing a dissimilarity metric when this pair of signatures is from the same person, and maximizing this dissimilarity metric when they belong to different persons. And the simple distance for approximating the "semantic" distance in the input space is obtained by mapping these two patterns using the same nonlinear sub-networks. Incorporated by deep learning, the Siamese architecture has been successfully applied to face recognition (Chopra et al., 2005), dimensionality reduction (Salakhutdinov & Hinton, 2007), and speech recognition (Chen & Salman, 2011). However, these Siamese neural networks are used for one single modality. The inputs to these networks are either images or speech representations.

Recent advances in multimodal deep learning have seen a trend to learn a joint representation by fusing different modalities (Ngiam et al., 2011; Srivastava & Salakhutdinov, 2012). (Man et al., 2012) suggests that information from different sensory channels converges somewhere in the brain to possibly form modality-invariant representations. Motivated by this, we propose a quasi-Siamese neural network for bimodal learning. Below we describe the details.

The quasi-Siamese has two sub-networks with the same architecture but different parameters. And these two networks are connected by some predefined compatibility measure. This network is shown in Figure 1. By designing a proper loss function from energy-based learning (Yann LeCun & Huang, 2006), we can learn the similar representations for these two bimodalities.

Formally, we denote the mapping from the inputs of these two sub-networks to the code layers as $f(p; W_f)$ and $g(q; W_g)$, in which, $f$ for image modality and $g$ for text modality; $W$ denotes the weight parameters in these two sub-networks. And the subscript in the weights $W$ denotes the corresponding modality. We define the compatibility measure between $i$th pair of image $p_i$ and the given tag words $q_i$ as

$$C(p_i, q_i; W_f, W_g) = \|f(p_i; W_f) - g(q_i; W_g)\|_1 \quad (9)$$

where $\|\cdot\|$ is the $\mathcal{L}_1$ norm.

To learn the similar representations of these two modalities for one object, we come up with a loss function given input $p_i$, $q_i$, and a binary indicator $\mathcal{I}$ with respect to the inputs, where $\mathcal{I} = 1$ if the tag words $q_i$ is for the image $q_i$, and $\mathcal{I} = 0$ otherwise. To simplify the notation we group the network parameters $W_f, W_g$ as $\Theta$. As the result, we define the loss function on any pair of inputs as

$$\begin{aligned} L(p_i, q_i, \mathcal{I}; \Theta) =& \alpha \left( L_I(p_i; \Theta) + L_T(q_i; \Theta) \right) \\ &+ (1-\alpha) L_C(p_i, q_i, \mathcal{I}; \Theta) \end{aligned} \quad (10)$$

where

$$L_I(p_i; \Theta) = \|p_i - \hat{p}_i\|_2^2 \quad (11a)$$
$$L_T(q_i; \Theta) = \|q_i - \hat{q}_i\|_2^2 \quad (11b)$$
$$L_C(p_i, q_i, \mathcal{I}; \Theta) = \mathcal{I} C^2 + (1-\mathcal{I}) \exp^{(-\lambda C)} \quad (11c)$$

Here, $\|\cdot\|_2$ is the $\mathcal{L}_2$ norm. $L_I$ and $L_T$ are the losses caused by data reconstruction errors for the given inputs (an image and its tag words) of two subnets. While $L_C(p_i, q_i, \mathcal{I}; \Theta)$ are contrastive losses incurred by whether the image and tag words are compatibility or not in two different situations indicated by $\mathcal{I}$. $\lambda$ in (11c) is a constant determinated by the upper bound of $C(p_i, q_i; \Theta)$ on all training data. $\alpha (0 < \alpha < 1)$ in the total loss function (11) is a parameter used to trade off between two groups of objectives, reconstruction losses and compatibility losses.

The learning for quasi-Siamese auto-encoder can be performed by standard back-propagation algorithm.

## 6. Choosing Alternatives

By obtaining the hierarchical three-level representations, the model is prepared for choosing alternatives. We have two strategies: a general strategy and a data-specific strategy. The general strategy is direct. To be specific, a pair of image $p_i$ and one of its tag words $q_i$ are taken into the network, the compatibility $L_C(p_i, q_i)$ between these two modalities can be calculated by equation (11c). Then, replacing the tag words $q_i$ with the other one $\widetilde{q}_i$, we can compute the other compatibility $L_C(p_i, \widetilde{q}_i)$ between the image $p_i$ and the other tag words $\widetilde{q_i}$. Finally, the tag words with larger compatibility are chosen as the correct tag for that image. Although this general strategy is applicable, we figure out another more accurate strategy by considering characteristics of data.

The data-specific strategy is based on one observation. To emphasize, for each image in GF the incorrect description is always the correct description of one other test image, which have been described in section 1. That means there exist loops among the tag words of some images. For example, three images $p_a$, $p_b$, and $p_c$, and their tag words form the tuples $(p_a, q_a, \widetilde{q_a})$, $(p_b, q_b, \widetilde{q_b})$, and $(p_c, q_c, \widetilde{q_c})$, respectively. There exist links among the six tag words of the three images. That is, for the six tag words we have either $\{\widetilde{q_a} = p_b, \widetilde{q_b} = p_c, \widetilde{q_c} = p_a\}$ or $\{\widetilde{q_a} = p_c, \widetilde{q_c} = p_b, \widetilde{q_b} = p_a\}$. And, once the tag words for an image is determined, the link can de resolved. We could find out the image with maximum discrepancy of compatibility between its two tag-words simply by

$$\arg\max_{i \in link} \left(L_C(p_i, q_i, \mathcal{I}) - L_C(p_i, \widetilde{q_i}, \mathcal{I})\right)^2 \quad (12)$$

for all images in one link.

To summary, we first find out all links in the test images. And then for each link we look for the most deterministic matching pair. Consequently, the set of images in the link are all resolved.

## 7. Experiments and Final Results

In this section, we report our experimental details and their results. In all our experiments, we only use the datasets ESP and GF provided by the organizer, though additional datasets can be used for training this model. Descriptions and some characteristics of the datasets ESP and GF have been given in section 1. We publish our implementation code at https://github.com/FangxiangFeng/deepnet, which is based on Nitish Srivastava's DeepNet library.

The ESP dataset has only the correct tag words for each image. Therefore, we need to generate an incorrect counterpart for word tags of each image in this dataset. This can be achieved by randomly choosing one from all the correct tag words of the rest images, while ensuring that each of the tag words occurs only one time.

In the training phase, the level-1, level-2, level-3 representations are extracted consecutively. The settings for learning level-1 representations has been described in section 3. In learning level-2 representations, we construct two stacked RBMs with the neurons configurations 1704-1024-1024 and 4000-1024-1024 for images and tag words, respectively. In learning level-3 representations, the quasi-Siamese auto-encoder with neurons configurations 1024-512-1024 both for images and tag words bimodal data, in which the free parameters $\alpha$ and $\lambda$ is set to 0.5 and 0.2, respectively. Additionally, we encourage the sparsity of the representations at all layers in the overall system.

Table 1. Public scores by different choosing strategies.

| Strategy | Score |
| --- | --- |
| Data-specific | 1.00000 |
| General | 0.87533 |

Table 2. The first five ranks on leaderboards.

| Team Name | Public Score | Private Score |
| --- | --- | --- |
| RBM | 1.00000 | 1.00000 |
| MMDL | 1.00000 | 1.00000 |
| BreakfastPirate | 1.00000 | 0.99158 |
| ryank | 1.00000 | 0.97983 |
| AuroraXie/ikretus | 0.72979 | 0.70507 |

We use both general and data-specific strategies described in section 6. To fulfill the computation of AUC (Area Under an ROC Curve), we express the dissimilarity of an image between its one tag words as a probability $P(p_i)$. We use Euclidean distances, though other metrics could be adopted. More specifically, $P(p_i)$ is expressed as

$$P(p_i) = \frac{(L_C(p_i, q_i, \mathcal{I}))^2}{(L_C(p_i, q_i, \mathcal{I}))^2 + (L_C(p_i, \widetilde{q_i}, \mathcal{I}))^2} \quad (13)$$

When the general strategy was adopted, we obtain the AUC with 0.87533; while the data-specific strategy can achieve the AUC with 100%, as in Table 1.

The public leaderboard show the scores achieved by all the 21 contesters. There are three teams also achieved score with 1.00000. And the fifth rank achieved score with 0.72979. The private leaderboard show the scores in the final test. The first two ranks teams RBM and MMDL achieved score with 1.00000. The first five ranks in leaderboard in public and private tests are listed in Table 2. Note that the last row in this table has two team names, the first for public test and the other for private test.

## 8. Discussion and Conclusions

Our results show that the solution is effective for this task. We believe that the strategy applied in choosing the alternatives is important. And in moderate representations are enough for make an accurate choice. For this reason, we did not tune the parameters very carefully and the learning cycles are reduced to speed up the overall learning process.

In conclusion, we construct a hierarchically bimodal representation and data-specific strategy for word tag alternative choice. These bimodal representations are obtained by three-stage extractions. In the first stage, the level-1 representations are achieved by extracting from images and texts using typical methods. In the second stage, the level-2 representations are learned by two consecutive RBMs for each modality. In the third stage, a quasi-Siamese auto-encoder is proposed for learning the level-3 representations. When choosing alternatives, we endeavor to find the maximum discrepancy among a link of images from an observation of the data characteristics.

## Acknowledgments

We thank the organizers for organizing this interesting competition. We also thank Nitish Srivastava for sharing his DeepNet library. Some part of the work was supported by National Sciences Foundation of China (No. 61273365) and the Fundamental Research Funds for the Central Universities (No. 2013RC0304).